\documentclass[runningheads]{llncs}
\usepackage{graphicx}

\begin{document}
\title{AdeNet: Deep learning architecture that identifies damaged electrical insulators in power lines}

\author{Ademola Okerinde \and Lior Shamir \and William Hsu \and Tom Theis}

\institute{Kansas State University, 2164 Engineering Hall, Manhattan, KS 43017-6221, USA  \\
\email{\{okerinde,lshamir,bhsu,theis\}@ksu.edu}}




\maketitle 

\begin{abstract}
Ceramic insulators are important to electronic systems, designed and installed to protect humans from the danger of high voltage electric current. However, insulators are not immortal, and natural deterioration can gradually damage them. Therefore, the condition of insulators must be continually monitored, which is normally done using UAVs. UAVs collect many images of insulators, and these images are then analyzed to identify those that are damaged. Here we describe AdeNet as a deep neural network designed to identify damaged insulators, and test multiple approaches to automatic analysis of the condition of insulators. Several deep neural networks were tested, as were shallow learning methods. The best results (88.8\%) were achieved using AdeNet without transfer learning. AdeNet also reduced the false negative rate to $\sim$7\%. While the method cannot fully replace human inspection, its high throughput can reduce the amount of labor required to monitor lines for damaged insulators and provide early warning to replace damaged insulators.

\end{abstract}




\section{Introduction}
\label{introduction}

Power line insulators change over time because they are continuously exposed to the weather, including heat, sun, and moisture. Deterioration can eventually damage insulators so they no longer function. Therefore identifying damaged insulators is a critical safety task. However, changes in characteristics such as color do not necessarily indicate that the insulator is nonfunctional. Therefore, identifying damaged insulators requires a diverse dataset of different types of damaged and undamaged insulators, at different ages and under different conditions. In this research, we explore various techniques to classify the condition of electrical insulators in power lines. 

Automatically detecting anomalies in image data has a broad range of applications \cite{beggel2019robust,golan2018deep}: automatically identifying product/equipment defects, medical diagnostics \cite{sidibe2017anomaly,wei2018anomaly}, quality checks at factories, and routine maintenance procedures. Models have been developed that can provide accurate image data. While most deep learning frameworks classify images well, detecting outliers and anomalies is more challenging. Training data for such tasks is often unbalanced because the number of anomaly samples, in many cases, is lower than the number of regular samples. 
Additionally, in some cases, the distinguishing characteristics of regular samples and anomalies are subtle; cracks, flashovers, scratches, and the like are difficult to identify. An effective anomaly detection model must, therefore, learn to identify these details as distinguishing characteristics. 


We developed a deep learning architecture, named AdeNet, to identify damaged power line insulators without pre-training. AdeNet also eliminates data augmentation overhead as discussed in \cite{malik2014comparative}. As a result, our model requires less computation and storage, making it more suitable for the computing capabilities of mobile/embedded devices such as unmanned aerial vehicles (UAV). Analysis on the device can be used to decide to acquire additional data while the UAV is still on its mission. AdeNet compares favorably with state-of-the-art deep learning (DL) architectures and shallow learning (SL) techniques. Our experimental results also show that the ROC score is not the best performance metric for an imbalanced dataset. A harmonic mean of recall and precision as expressed by the F1 score is a better choice. 


\section{Related Work}
\label{related_work}

Detecting anomalies or outliers in image data is important research for computer vision. This task is often characterized by a number of challenges, so it differs from typical supervised machine learning. One of these challenges is that anomaly samples are far fewer than regular samples. 

Many methods have been proposed to address this challenge at both the data and algorithm levels. In \cite{bergmann2019mvtec}, a comprehensive real-world dataset is curated for unsupervised detection of anomalies. A common approach is to train the classifier solely on the normal samples with the hope that the architecture will be robust enough to capture the intrinsic characteristic features of the normal class and thus can identify abnormal characteristics by inference \cite{rudolph2020same}. Notable state-of-the-art methods include AnoGAN \cite{schlegl2017unsupervised,schlegl2019f}, L2 and SSIM Autoencoder \cite{bergmann2018improving,bergmann2019mvtec}, CNN Feature Dictionary \cite{napoletano2018anomaly}, GMM-Based Texture Inspection Model \cite{bottger2016real}, and Variation Autoencoder. 


Because untreated damaged insulators are inherently dangerous, automatic monitoring of power line insulators have been substantially researched \cite{jenssen2018automatic,miralles2014state}. A basic approach for detecting insulators in aerial images used morphology analysis with Otsu threshold followed by a support vector machine (SVM) classifier \cite{wang2016insulator}. Another approach detected edges and corners represented through MultiScale MultiFeature descriptors \cite{liao2014robust}. Lattice detection has also been tested with good results in detecting anomalies in insulators in power lines \cite{zhao2012detecting}.

In the specific context of automatically detecting damaged insulators, Structural Similarity \cite{wang2004image} can determine the degree of distortion in damaged insulators from undamaged ones. However, as shown in our experiment, this approach works well if the distortion is digital or artificial, not the result of natural degradation. Circular GLOH-Like descriptors were used for detecting and classifying insulators \cite{miao2019insulator}. A simple method based on shape and the distribution of brightness was proposed as a way to automatically identify cables damaged by lightning  \cite{ishino2004detection}. Common deep neural network architectures such as SSD \cite{liu2016ssd} and ResNet \cite{he2016deep} have also been used to identify damaged power lines in aerial images \cite{jenssen2018automatic}.

\cite{miao2019insulator} proposed an effective and reliable method of detecting anomalies in insulators using a deep learning technique for aerial images. In the proposed deep detection approach, the single shot multibox detector (SSD), a powerful deep meta-architecture, using two-stage fine-tuning could automatically extract multi-level features from aerial images, thus obviating manual extraction. Inspired by transfer learning, a two-stage fine-tuning strategy was implemented with separate training sets. In the first stage, the basic insulator model was obtained by fine-tuning the COCO model with aerial images, including different types of insulators and various backgrounds. In the second stage, the basic model was fine-tuned using training sets with specific insulator types and specific situations. After the two-stage fine-tuning, the well-trained SSD model can directly and accurately identify the insulator in aerial images. 

According to \cite{liu2016method}, the insulator is important to a transmission line, and detecting defects in insulators relies heavily on insulator position. Traditional methods of insulator recognition depend on color features and geometric features. These methods are influenced illumination and background, among other things, resulting in poor generalization. \cite{liu2016method} proposed a method where insulators were recognized using deep learning algorithms. First, \cite{liu2016method} constructed a training dataset that had three categories: insulator, background, and tower. Second, \cite{liu2016method} initialized the convolution neural networks as a six-level network and adjusted training parameters to train the model. Finally, the trained model was used to predict the candidate insulator position. With the help of a non-maximum suppression algorithm and the line fitting method, \cite{liu2016method} identified the exact location of an insulator.



\section{Proposed Approach}
\label{approach}

Paradigms of automatic image classification can be broadly divided into DL and SL. DL often requires more difficult training and a large training set, but it has the advantage of being non-parametric. It is often achieves more accurate than SL. SL requires fewer training samples and is normally less prone to overfitting. However, image features from each image can be computationally expensive, and the resulting machine learning model is not always as accurate as models based on deep learning. In this research, we attempted both DL and SL to compare the two approaches to automatically detecting damaged power line insulators.


\subsection{Shallow Learning Approach}

For shallow learning, we tested the Udat shallow learning image analysis tool \cite{shamir2017udat}, which implements the Wndchrm method \cite{shamir2008wndchrm,orlov2008wnd}. Udat works by first extracting a large set of 2841 numerical image content descriptors from raw pixels and transforming the raw pixels \cite{shamir2008evaluation}. The numerical image content descriptors are shape, edges, textures, polynomial approximation, statistical distribution of pixel values, and others for a comprehensive numerical representation of the image \cite{shamir2010impressionism}. Detailed information about the Wndchrm algorithm can be found in \cite{shamir2008wndchrm,orlov2008wnd,shamir2008evaluation,shamir2010impressionism}.

Once the numerical image content descriptors are computed for all images, the values are then used for mature classification algorithms like Random Forest, Support Vector Machine, Gradient Boosting, and Naive Bayes.





\subsection{Deep Learning Approach}

In deep learning, multiple layers of mini-algorithms, called neurons, work together to draw complex conclusions. We chose the AdeNet architecture and compared it with several deep neural network architectures, from the basic LeNet-5 to more complex ResNet-101, VGG19, and MobileNetV2. 

\subsubsection{MobileNetV2}

MobileNetV2 \cite{sandler2018mobilenetv2} is a mobile architecture based on Convolutional Neural Networks (CNNs). The basic building block is a bottleneck depth-separable convolution with residuals. Our implementation of MobileNetV2, taken from tensorflow-keras applications, has 398, 690 trainable, and 14,000 non-trainable parameters. The total size of the model is 5.7MB. 

\subsubsection{AdeNet}
\label{adenet}

\begin{figure}
\centering
\includegraphics[scale=0.30]{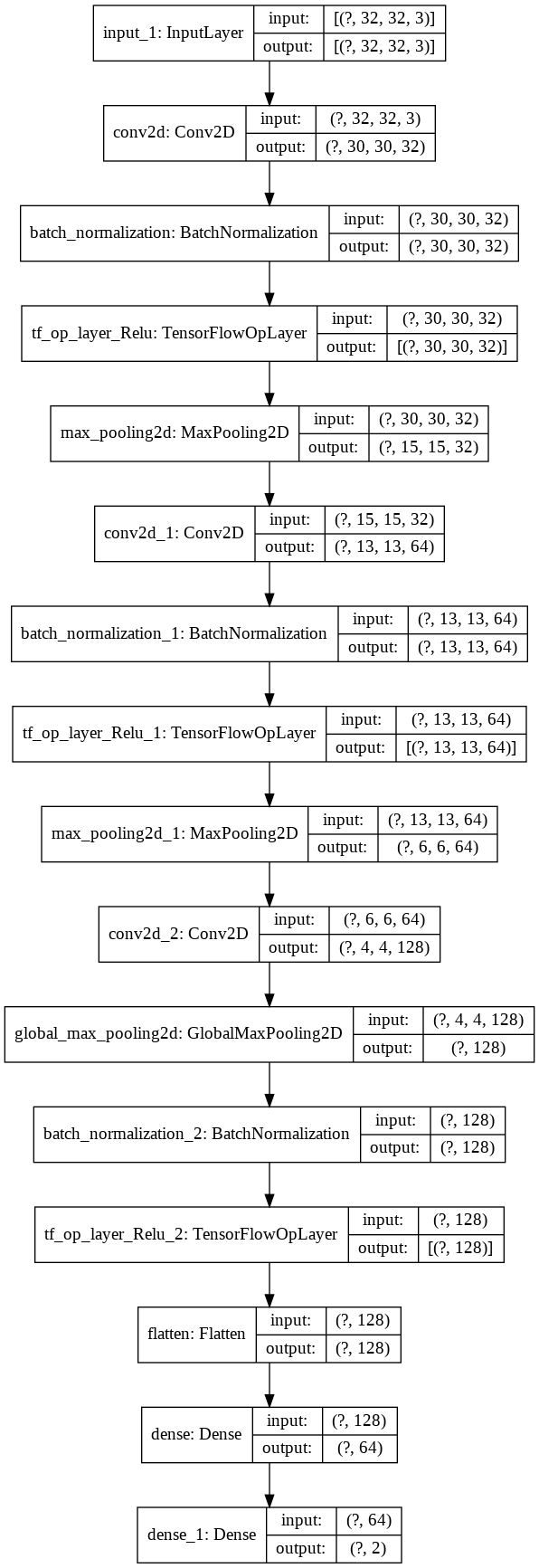}
\caption{AdeNet Deep Learning Architecture}
\label{fig:adenet}
\end{figure}

AdeNet is a deep learning architecture implemented with three layers of CNNs, each with batch normalization, maxpooling, and ReLU with no dropout. One fully connected layer comes before the softmax layer. Figure \ref{fig:adenet} shows the architecture. It contains the initial fully convolutional layer with 32 filters. We always used kernel size 3$\times$3 as is standard for modern networks. The trained model size was 1.3MB with 102,082 trainable parameters and 448 non-trainable parameters.


\section{Methodology and Experimental Design}

\subsection{Dataset}
\label{Dataset}

The dataset was provided by Black and Veatch, consisting of 1696 JPEG images of power lines  with the dimensions of 5280$\times$3956 and a resolution of 72x72. We randomly divided the entire dataset into 80\% training and 20\% test sets. Each image is annotated with a bounding-box around the objects of interest: the insulators. The insulators are broadly categorized as damaged (positive class) and undamaged (negative class). Table~\ref{tab:dataset} shows the number of samples in each class after cropping out the insulators.

\begin{table}[ht]
  \caption{Number of samples of damaged and undamaged insulators in the dataset.}
  \label{tab:dataset}
  \begin{tabular}{lccc}
  \hline
Dataset  & Damaged & Undamaged & \#images\\
\hline
Train + Validation  & 1417 & 2836 & 1484\\
Test  & 290 & 835 & 212\\
  \hline
  \end{tabular}
\end{table}

The dataset presents a number of interesting challenges: within class variations, class imbalance, noisy background, and varying orientations and scales. Using within class variations as an example, in our experiments, we classified as damaged any insulator with flash-over or that was broken, missing, or fried. This leads to skewness of samples within each class as well.

\subsection{Data Pre-processing}

\begin{figure}
  \centering
  \includegraphics[width=\linewidth]{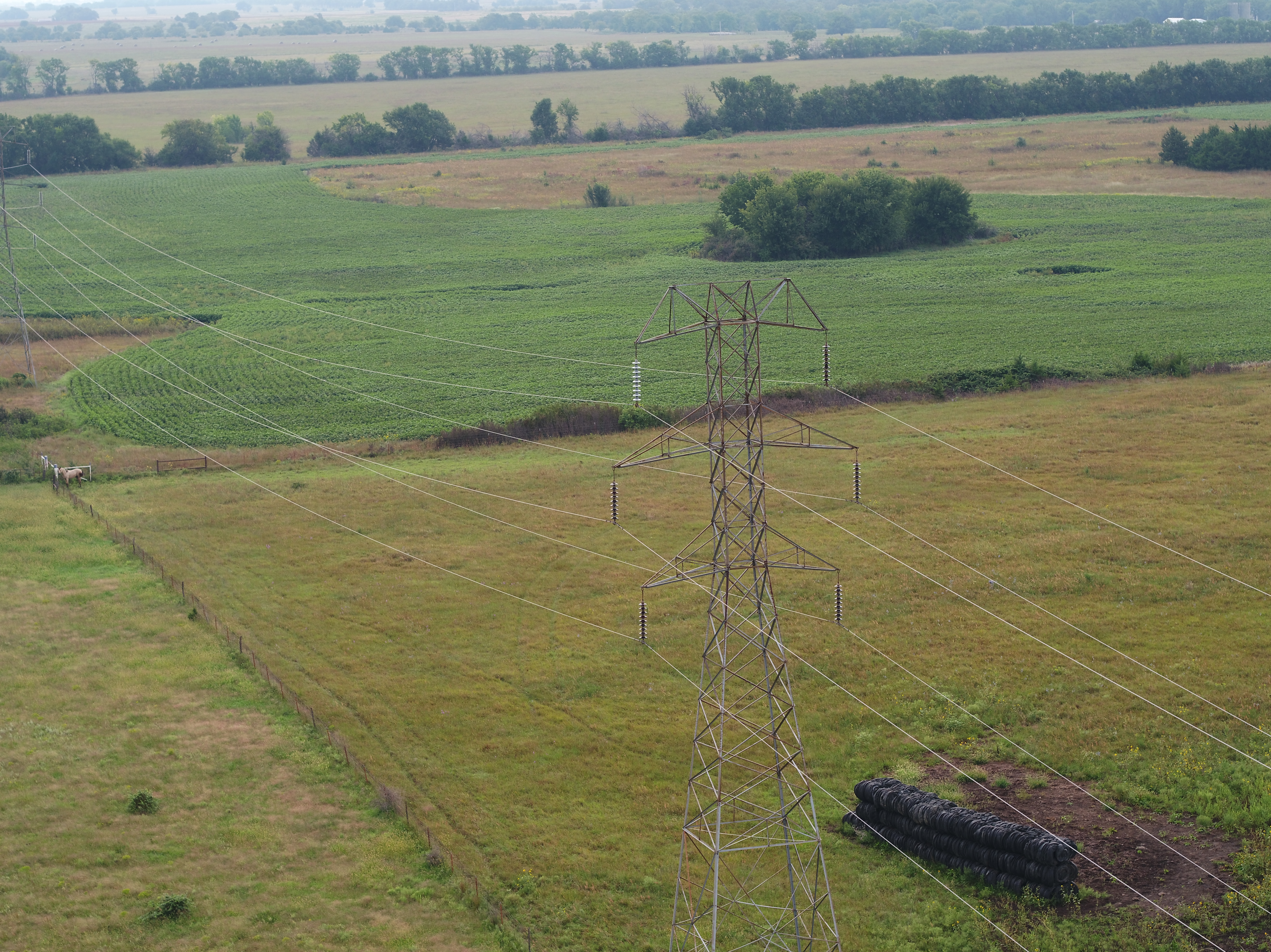}
  \caption{An example of a power line image.}
  \label{fig:powerline}
\end{figure}

\begin{figure}
    \centering
    \includegraphics[width=.1\textwidth]{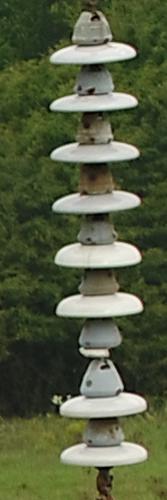}\hfill
    \includegraphics[width=.1\textwidth]{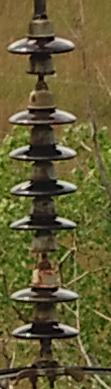}\hfill
    \includegraphics[width=.1\textwidth]{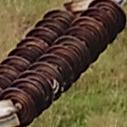}\hfill
    \includegraphics[width=.1\textwidth]{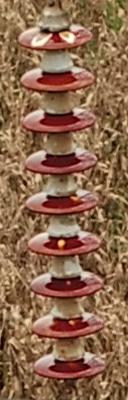}\hfill
    
    \includegraphics[width=.1\textwidth]{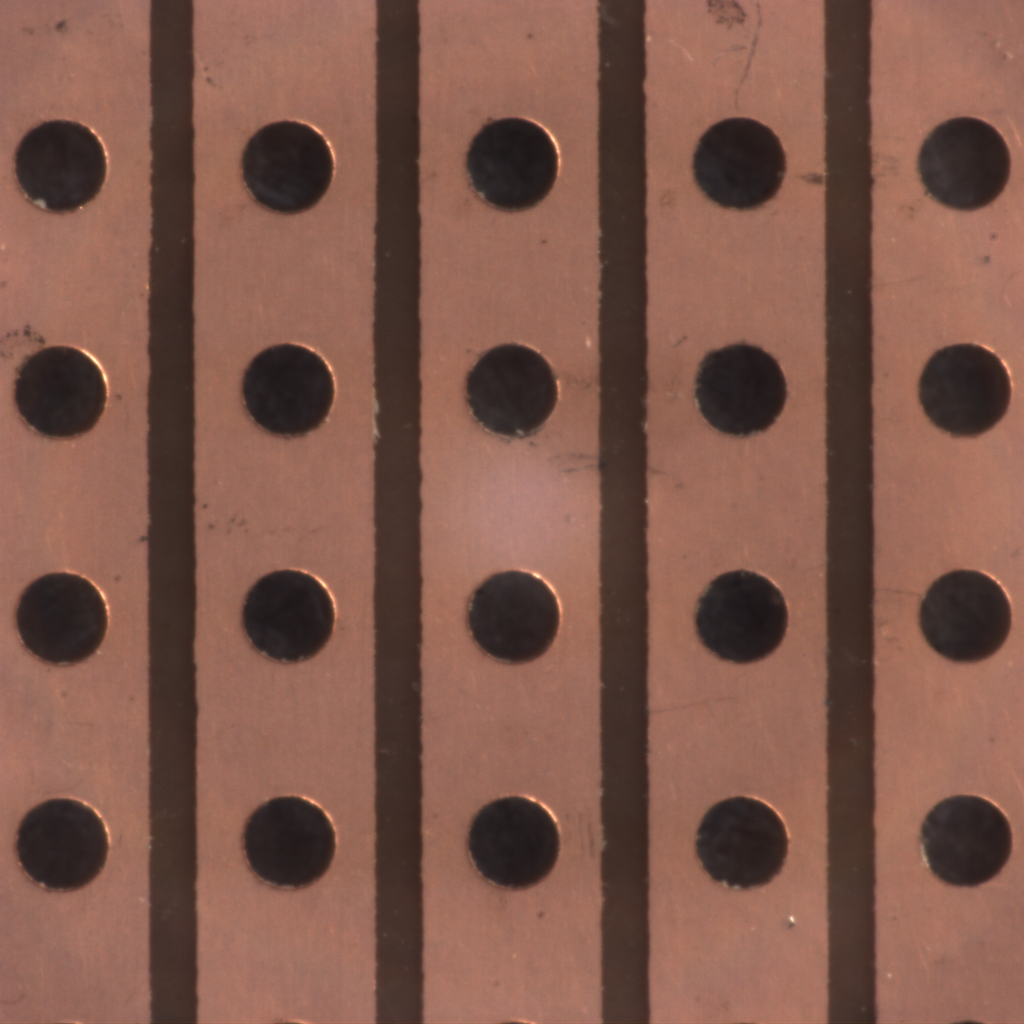}\hfill
    \includegraphics[width=.1\textwidth]{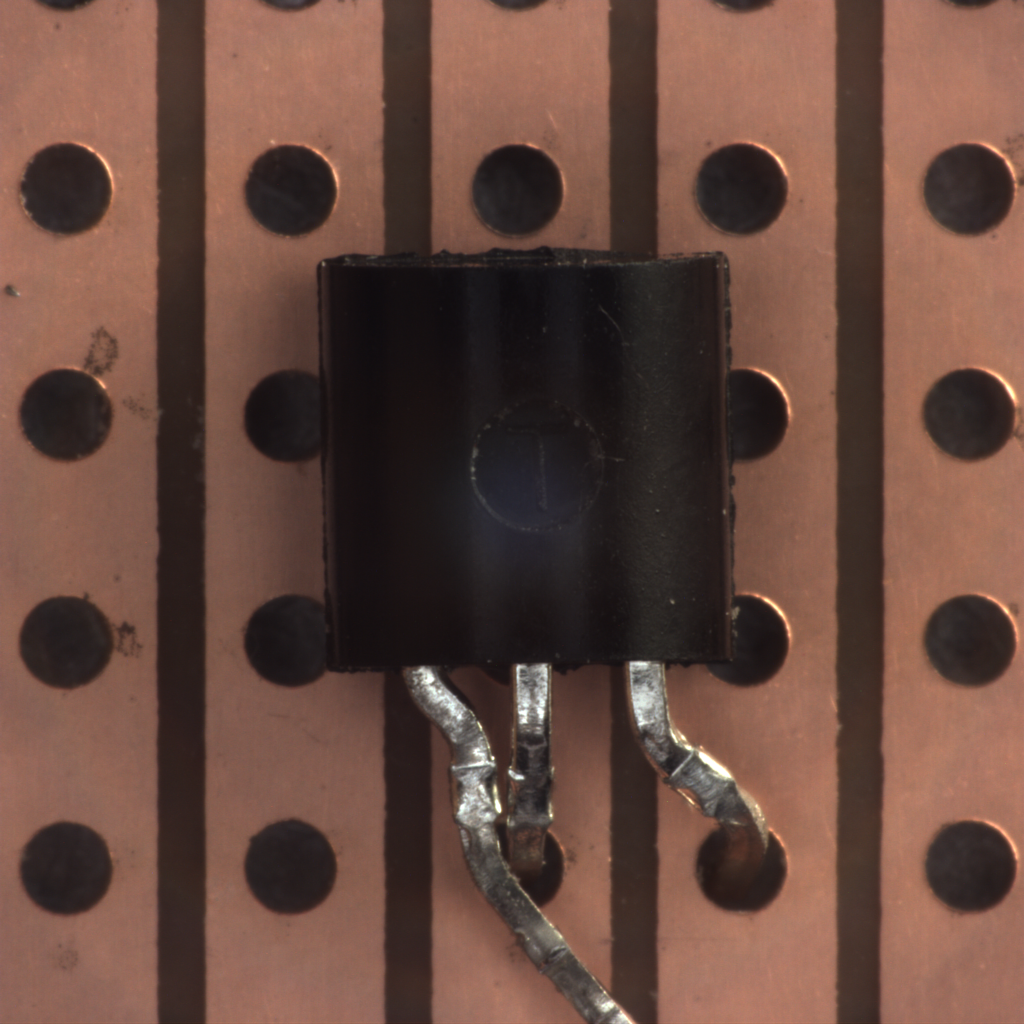}\hfill
    \includegraphics[width=.1\textwidth]{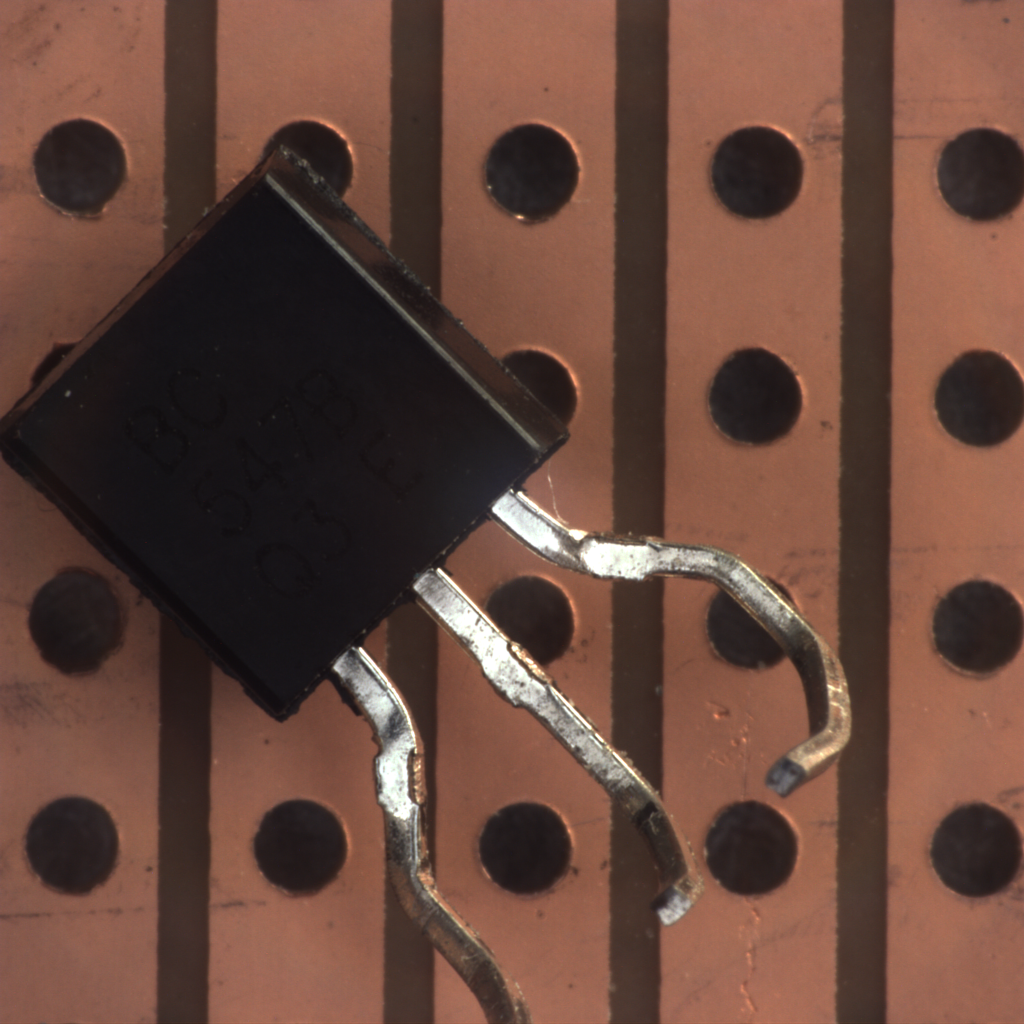}\hfill
    \includegraphics[width=.1\textwidth]{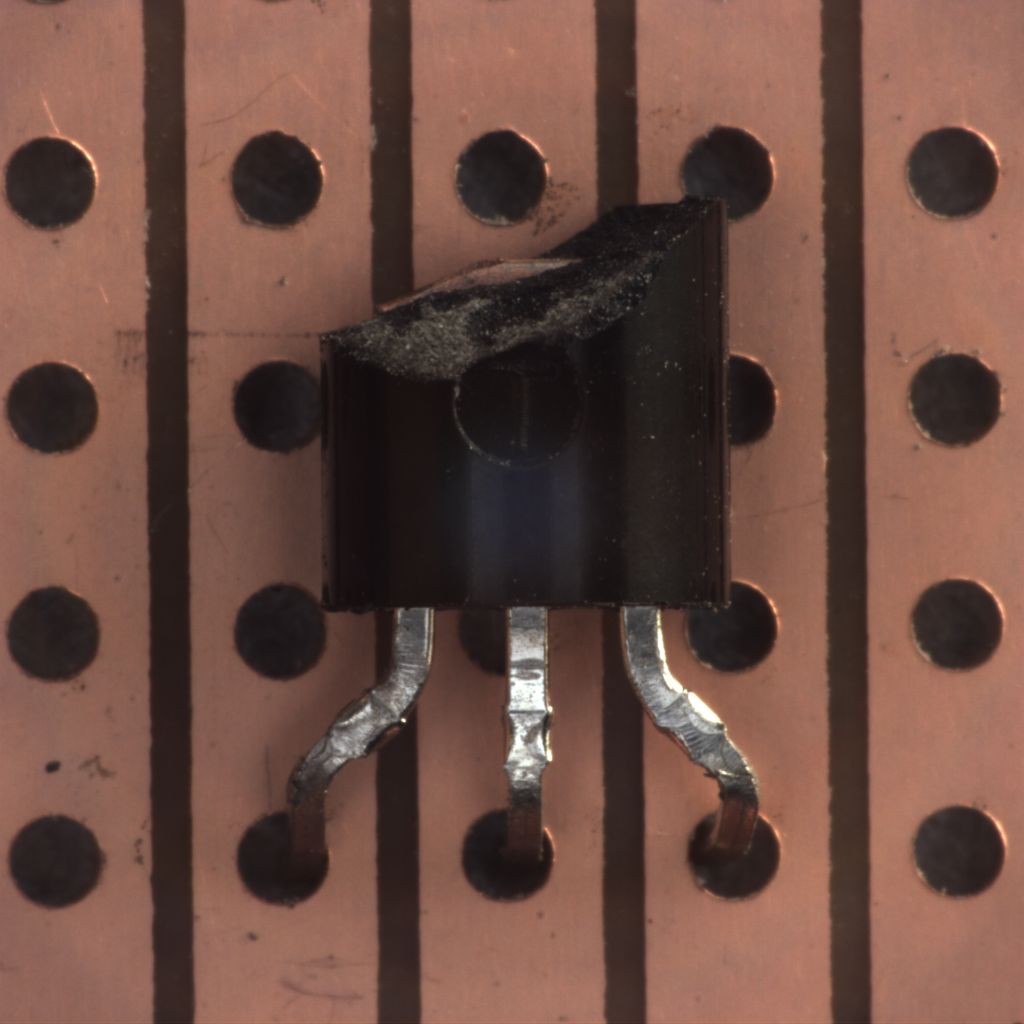}\hfill
    \\[\smallskipamount]
    \caption{Damaged insulators (top) and transistors (bottom)}
    \label{fig:insulators}
\end{figure}

\begin{figure}
    \centering
    \includegraphics[width=.1\textwidth]{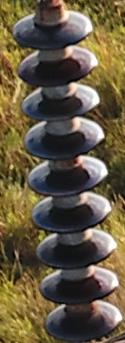}\hfill
    \includegraphics[width=.1\textwidth]{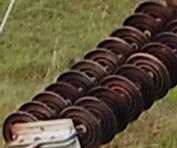}\hfill
    \includegraphics[width=.1\textwidth]{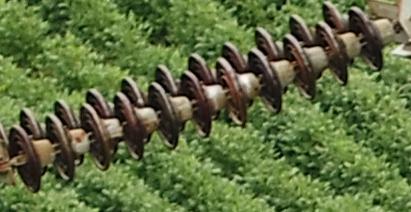}\hfill
    \includegraphics[width=.1\textwidth]{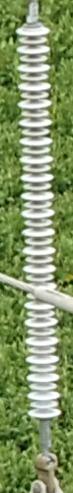}\hfill
    
    \includegraphics[width=.1\textwidth]{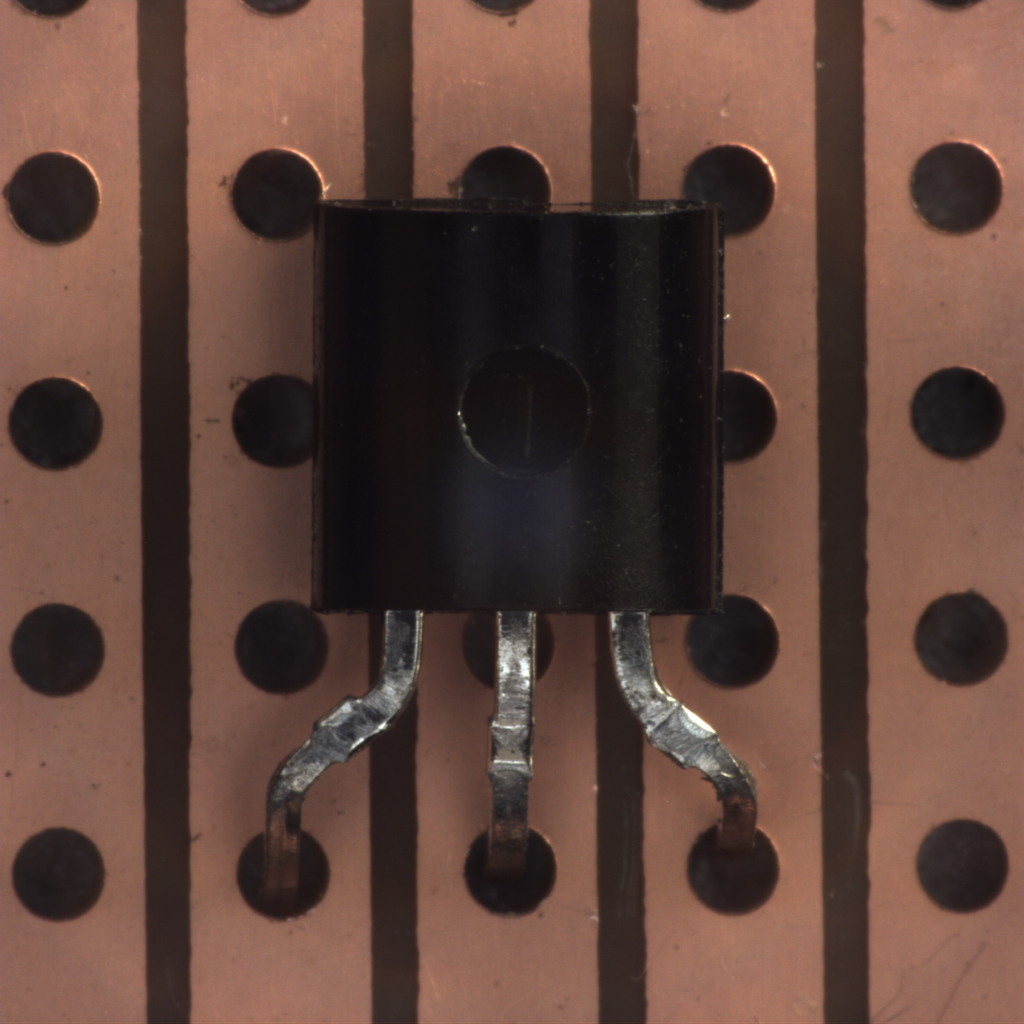}\hfill
    \includegraphics[width=.1\textwidth]{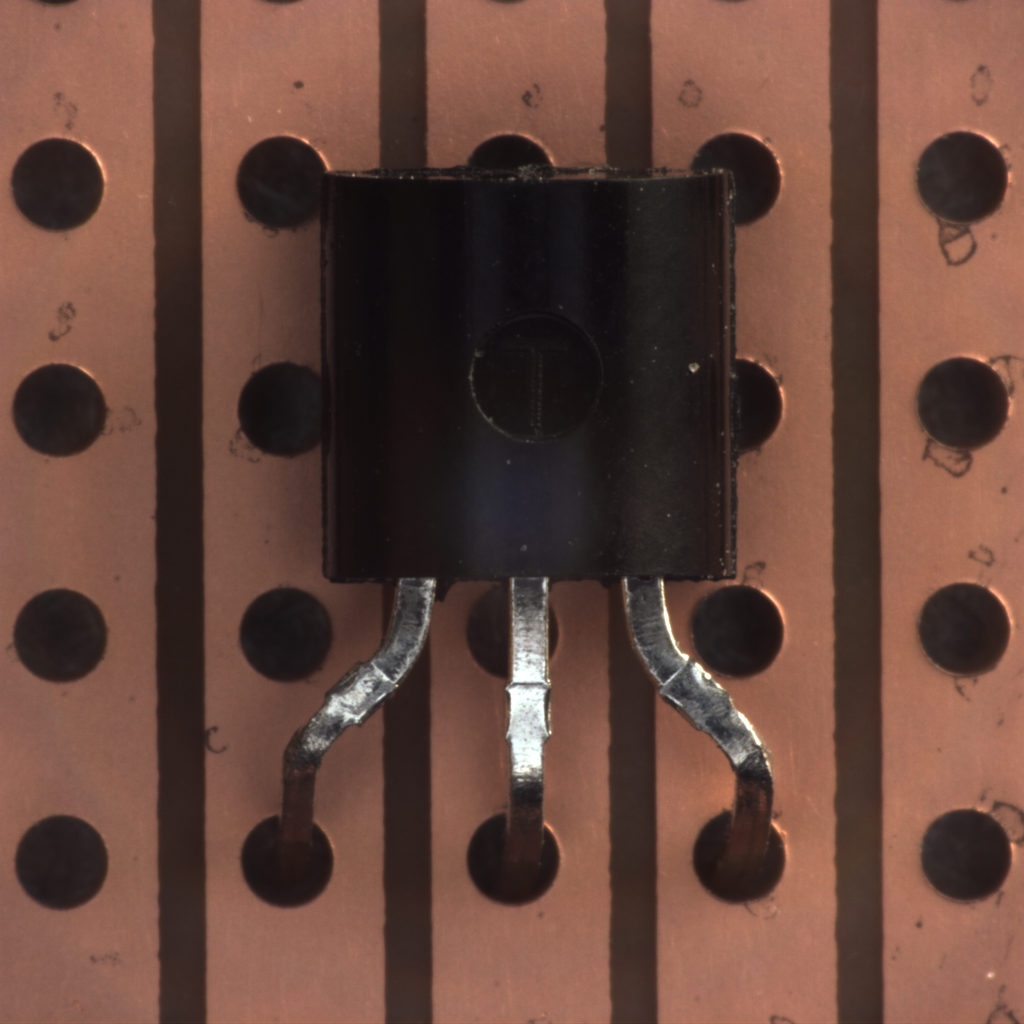}\hfill
    \includegraphics[width=.1\textwidth]{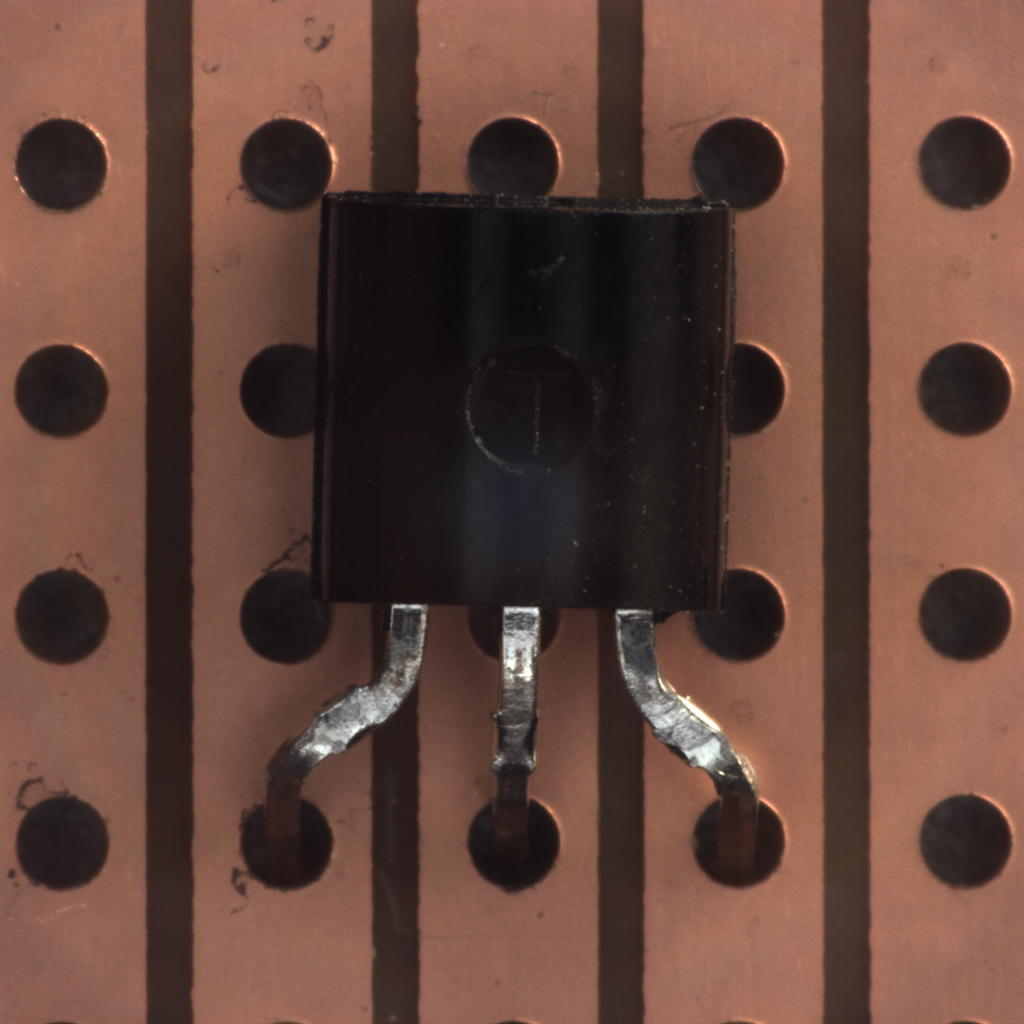}\hfill
    \includegraphics[width=.1\textwidth]{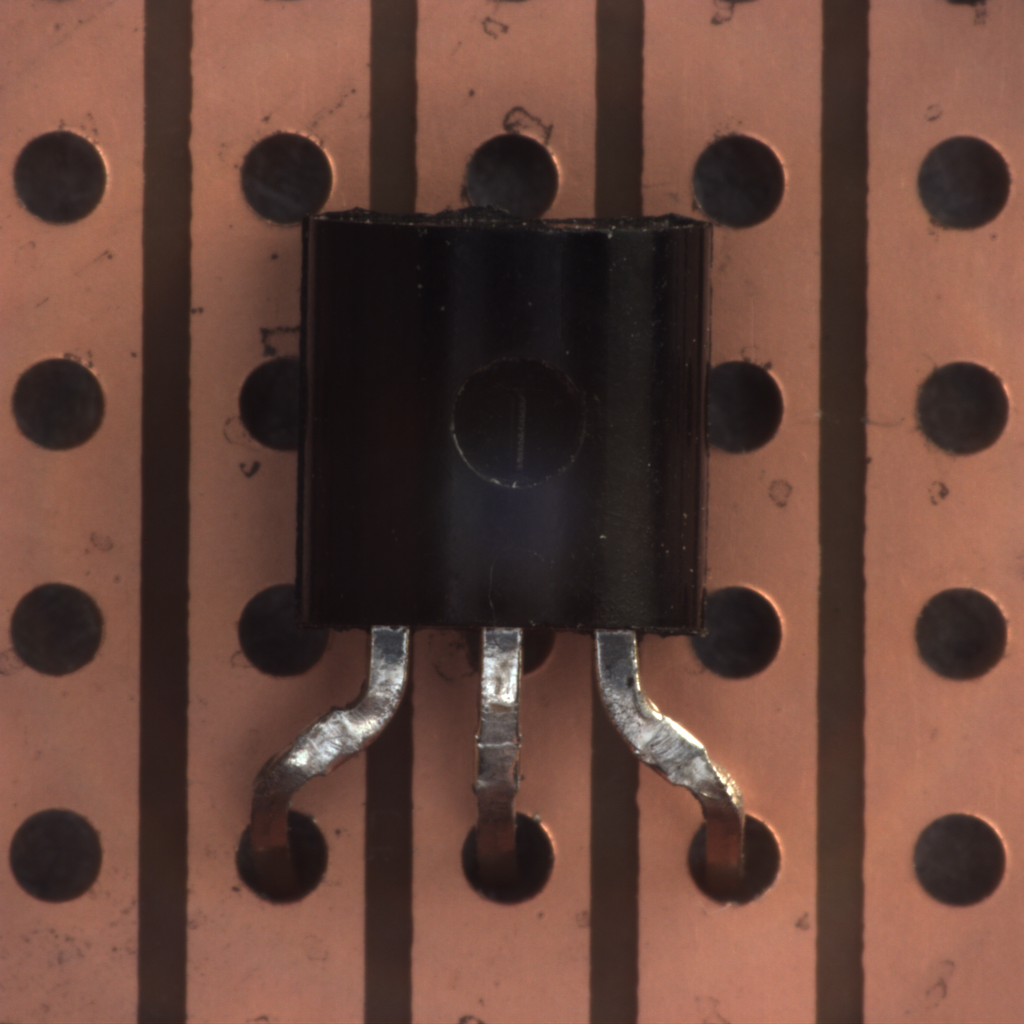}\hfill
    \\[\smallskipamount]
    \caption{Undamaged insulators (top) and transistors (bottom)}
    \label{fig:insulators_undamaged}
\end{figure}

Figure~\ref{fig:powerline} shows an example of a power line image. The electrical insulators were then cropped out of images for feature extraction, giving us a total of 4253 images for training and validation as well as 1125 images for testing. Figure~\ref{fig:insulators} shows samples of damaged insulators, and figure~\ref{fig:insulators_undamaged} shows samples of undamaged insulators. For SL, we used Udat to extract the 2900 most informative image features from the insulators. Udat uses statistical techniques like Radon transform features,  Chebyshev Statistics, and Multi-scale Histograms for image feature extraction. For DL, because insulators varied in scale, we padded them to the max dimension in each batch during training and testing.



\begin{figure}
    \centering
    \includegraphics[width=.4\textwidth]{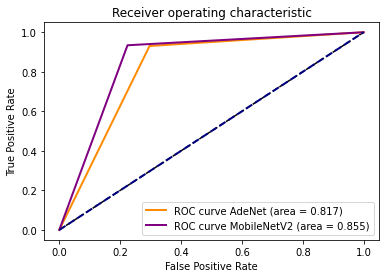}\hfill
  
    \caption{Model comparison of the ROC curve on insulator dataset}
    \label{fig:roc_curve}
\end{figure}


\begin{figure}
    \centering
    \includegraphics[width=.2\textwidth]{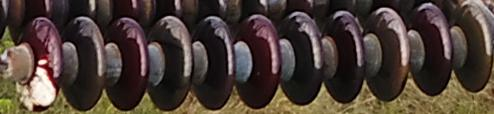}
    \includegraphics[width=.2\textwidth]{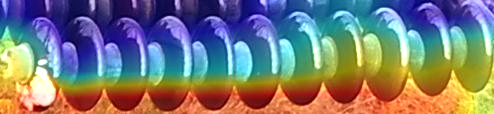}
    \includegraphics[width=.1\textwidth]{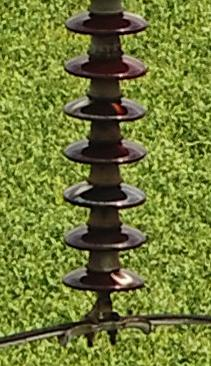}
    \includegraphics[width=.1\textwidth]{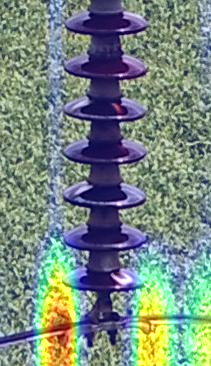}
    \\
    \includegraphics[width=.1\textwidth]{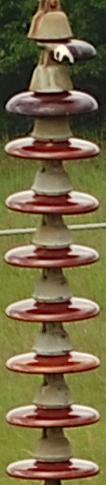}
    \includegraphics[width=.1\textwidth]{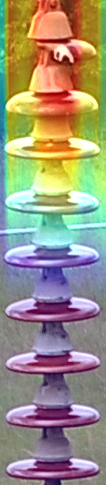}
    \includegraphics[width=.1\textwidth]{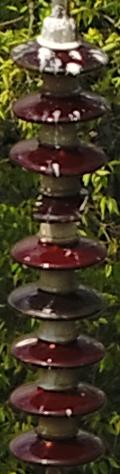}
    \includegraphics[width=.1\textwidth]{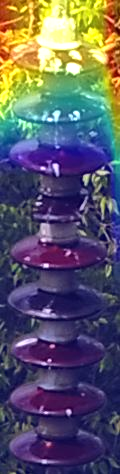}
    \includegraphics[width=.1\textwidth]{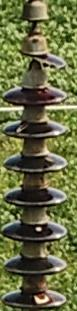}
    \includegraphics[width=.1\textwidth]{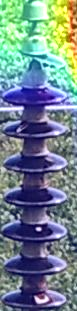}
    \\[\smallskipamount]
    \caption{Grad-CAM heatmap visualizations}
    \label{fig:visualization}
\end{figure}









\section{Results and Discussion}


The different methods were applied to the dataset described in Section~\ref{Dataset}. For training and testing, we used mostly 5-fold as the different classifiers for cross-validation. Each split was trained with 10 or 20 epochs. For transfer learning, we used models pre-trained on the ImageNet dataset. We used a batch size of 16 for training and testing. Most of our DL experiments were run on Tesla P100-PCIE on Google Colab. The DL models were built using TensorFlow \cite{abadi2016tensorflow}. 

The results are shown in tables~\ref{tab:experiments_val} and \ref{tab:experiments}. For the analysis of the results, the table gives accuracy, F1 score, precision, recall, and ROC area under the curve for each classifier. The measurements are the average of all folds.

We found that MobileNetV2 sometimes performed better if trained using more epochs, starting with saved model weights. The performance often dropped if the F1 score for the initial 10 epochs reached 91\% in the damaged class. The model often inverted performance from one class to favor another if the results for the first 10 epochs were skewed. As tables~\ref{tab:MV2} and \ref{tab:MV2-2} show, the model had fewer false negatives when trained for fewer epochs. This is crucial for this situation.

Without transfer learning, MobileNetV2 was highly influenced by the class imbalance because most data were classified as negatives. Likewise, LeNet-5 \cite{lecun2015lenet}, EfficientNetB7 \cite{tan2019efficientnet}, VGG19 \cite{simonyan2014very}, and ResNet-101 \cite{he2016deep} models all classified most data as negatives. All models other than LeNet-5 were pre-trained on the ImageNet dataset \cite{deng2009imagenet}. Table~\ref{tab:experiments_val} shows their performance on different validation percentage splits.

Our experimental findings showed that all MobileNetV2 model performance on the test set correlated with results from the validation set. For example, if a model did well on the positive class in validation, it also performed well on the positive class in the test.
We ran all the five models on the test set, and the results showed that MobileNetV2 performed better on 20-epochs-models on the test set. We reported the average of the five models in Table~\ref{tab:experiments}.

Comparing AdeNet with similar simple DL architectures like LeNet5 revealed that an overly simple network was not optimal, but a carefully engineered simple architecture that balanced between simple and complex designs, like ResNet-101 and VGG19, worked. The confusion matrices in tables~\ref{tab:classifier}, \ref{tab:MV2}, \ref{tab:MV2-2}, \ref{tab:adenet_1} and \ref{tab:adenet_2} 
reveal that AdeNet was significantly better at classifying both classes of the dataset. This was confirmed by  high F1 scores in table~\ref{tab:experiments}. F1 score is a better performance metric for evaluating models developed for imbalanced datasets. Consider, for instance, random forest with high ROC
score, but poor F1 score (see table~\ref{tab:classifier}).

Due to small number of samples in \cite{bergmann2019mvtec}'s transistor category, AdeNet performed poorly with and without fine-tuning because the entire dataset was classified as negatives, i.e., undamaged transistors. Borrowing intuition from \cite{wang2020frustratingly}, we experimented with freezing two, ten, and thirteen layers of pre-trained AdeNet on the insulator dataset to see how well it transferred to the transistor detection task. This reinforced the conclusion that a DL model performs better with more data. However, SL with Udat achieved an accuracy of 92.8\% (10-fold-cross-validation split) on validation data with roughly half of the damaged class miss-classified.

With Grad-CAM, we used heatmap to visualize the model's attention on the insulator while making predictions as shown in Figure \ref{fig:visualization}. Grad-CAM uses the penultimate (pre-dense) convolutional layer containing spatial information that is completely lost in dense layers. The model's reasoning did make some sense from the human visual system perspective. The area that constituted damage in the insulators were located where the heatmap was most intense, i.e., heavily red.

Figure~\ref{fig:roc_curve} compares the ROC curve for the models using the Black and Veatch insulator dataset. The curve shows that MobileNetV2 was slightly better than AdeNet. This plot was based on one model. Table~\ref{tab:experiments} shows that, on average, AdeNet outperformed MobileNetV2.

A major hyper-parameter tuning task was to determine how long to train a deep neural network. We experimented with using early stopping and fixed epochs, finding that performance was much the same. However, with callbacks, training time was significantly reduced.

Of the tested classifiers, the best shallow learning classifier was Random Forest. However, its confusion matrix (see Table~\ref{tab:classifier}) revealed that the classifier did not do well in classifying the damaged class, making it the worst classifier for this task.

\begin{table}[ht]
  \caption{Confusion matrix for Random Forest on test set}
  \label{tab:classifier}
  \begin{tabular}{lcc}
& Predicted Damaged & Predicted Undamaged \\
\hline
Actually Damaged & 11 & 279  \\
Actually Undamaged & 2 & 833  \\
  \hline
  \end{tabular}
\end{table}

\begin{table}[ht]
  \caption{Confusion matrix for MobileNetV2 after 20 epochs for test set}
  \label{tab:MV2}
  \begin{tabular}{lcc}
& Predicted Damaged & Predicted Undamaged \\ 
\hline
Actually Damaged & 991 & 459  \\ 
Actually Undamaged & 1063 & 3112 \\ 
  \end{tabular}
\end{table}

\begin{table}[ht]
  \caption{Confusion matrix for MobileNetV2 after 10 epochs for test set}
  \label{tab:MV2-2}
  \begin{tabular}{lcc}
& Predicted Damaged & Predicted Undamaged \\
\hline
Actually Damaged & 724 & 726 \\
Actually Undamaged & 312 & 3863 \\
  \hline
  \end{tabular}
\end{table}

\begin{table}[ht]
  \caption{Confusion matrix for AdeNet after 10 epochs for test set}
  \label{tab:adenet_1}
  \begin{tabular}{lcc}
& Predicted Damaged & Predicted Undamaged \\ 
\hline
Actually Damaged & 947 & 503  \\ 
Actually Undamaged & 257 & 3918 \\ 
  \hline
  \end{tabular}
\end{table}

\begin{table}[ht]
  \caption{Confusion matrix for AdeNet after 20 epochs for test set}
  \label{tab:adenet_2}
  \begin{tabular}{lcc}
& Predicted Damaged & Predicted Undamaged \\ 
\hline
Actually Damaged & 1026 & 424  \\ 
Actually Undamaged & 213 & 3962 \\ 
  \hline
  \end{tabular}
\end{table}

\subsection{Ablation Studies}
We inspected parts of the architecture and parameters for analysis. 

\textbf{Batch Normalization} We observed no significant differences in model performance with or without batch normalization \cite{ioffe2015batch} possibly because AdeNet is not exceptionally deep. We could thus further reduce the parameter by $\sim$1\%.

\section{CONCLUSION}

In this research, we addressed the task of detecting damaged electrical insulators in power line images, an important safety task. Detecting damaged electrical insulators has been a labor-intensive task requiring substantial experience and careful examination of the images. 

The new architecture, AdeNet, based on a deep convolutional neural network has the advantage of requiring little energy, allowing it to be used on low-energy devices like UAVs, which are often used to acquire the data for identifying damaged insulators. The low energy allows AdeNet to make decision on the UAV, which can then be used to acquire more data of suspect insulators during a mission, without going back to a site and re-using the UAV. Like intuition in \cite{hacohen2019power,frankle2018lottery}, we found an architecture that is not as simple as LeNet5 and not as complex as ResNet (just the right mix of simplicity and complexity) that also performs well in detecting anomalies in damaged insulators in power lines. 

In comparing the proposed method and other solutions, including shallow learning, we found that deep learning outperforms shallow learning architectures. Within DL architecture, fine-tuning the pre-trained CNN models did not improve the performance in all cases. We demonstrated this through experiments with our AdeNet architecture. Experimental results also showed that our architecture, without pre-training, outperformed all other deep neural networks. This makes AdeNet suitable for mobile/embedded devices that have storage and computation constraints.

\begin{table*}[ht]
  \caption{Classification accuracy, F1, precision, recall, and ROC area under the curve when using different methods on varying validation set.}
  \label{tab:experiments_val}
  \begin{tabular}{lccccccc}
  \hline
Dataset   & Classifiers & Acc & F1 & Precision & Recall & ROC Area & \#folds \\
\hline
val & Random Forest   & 0.86 & 0.85 & 0.86 & 0.86 & 0.92 & 10 \\
val & MultiLayer Perceptron  & 0.82 & 0.82 & 0.82 & 0.82 & 0.86 & 10 \\
val & Gradient Boosting  & 0.87 & - & - & - & - \\
\hline
val & LeNet-5  & 0.68 & 0.46 & 0.70 & 0.53 & 0.53 & 5 \\
val & AdeNet + 10 epochs & 0.86 & 0.83 & 0.87 & 0.81 & 0.81 & 5\\ 
val & EfficientNetB7  & 0.63 & 0.36 & 0.36 & 0.49 & 0.49 & 5\\
val & VGG19  & 0.67 & 0.40 & 0.44 & 0.50 & 0.50 & 5 \\
val & ResNet-101  & 0.59 & 0.40 & 0.35 & 0.51 & 0.51 & 5\\

  \hline
  \end{tabular}
\end{table*}

\begin{table*}[ht]
  \caption{Classification accuracy, F1, precision, recall, and ROC area under the curve when using different methods on test data (average of 5 folds cross validation).}
  \label{tab:experiments}
  \begin{tabular}{lcccccc}
  \hline
Learning   & Classifiers & Acc & F1 & Precision & Recall & ROC Area  \\
\hline
Shallow & Random Forest   & 0.79 & 0.64 & 0.79 & 0.62 & 0.82  \\
Shallow & Random Tree   & 0.69 & 0.60 & 0.60 & 0.60 & 0.60  \\
Shallow & Naive Bayes   & 0.56 & 0.56 & 0.64 & 0.67 & 0.76  \\
Shallow & MultiLayer Perceptron  & 0.77 & 0.69 & 0.69 & 0.68 & 0.79  \\
Shallow & Support Vector Machine & 0.72 & 0.65 & 0.65 & 0.66 & 0.66 \\
\hline
Deep & MobileNetV2 + 10 epochs  & 0.82 & 0.72 & 0.77 & 0.71 & 0.73  \\
Deep & MobileNetV2 + 20 epochs & 0.70 & 0.63 & 0.76 & 0.70 & 0.71  \\
Deep & AdeNet + 10 epochs  & 0.86 & 0.81 & 0.85 & 0.80 & 0.80 \\ 
{\bf Deep} & {\bf AdeNet} + {\bf 20 epochs}  & {\bf 0.89} & {\bf 0.84} & {\bf 0.87} & {\bf 0.83} & {\bf 0.83}  \\

  \hline
  \end{tabular}
\end{table*}

\section*{Acknowledgement}
 We are grateful to Kansas State University Beocat team for providing us with access to the high performance computing cluster for running the experiments.


\bibliographystyle{splncs04}
\bibliography{references}

\end{document}